\documentclass[10pt,twocolumn,letterpaper]{article} 

\usepackage{avss}
\usepackage{times}
\usepackage{epsfig}
\usepackage{graphicx}
\usepackage{amsmath}
\usepackage{amssymb}
\usepackage{amsmath,epsfig}
\usepackage{xspace}
\usepackage{amssymb, amsmath}
\usepackage[T1]{fontenc}
\usepackage{cite}
\usepackage{orcidlink}
\usepackage{academicons}
\usepackage{graphicx}
\usepackage{booktabs}
\usepackage{nicematrix}
\usepackage{adjustbox}
\usepackage{float}
\usepackage{tabularx}
\usepackage{multicol,multirow,array}
\usepackage{academicons}
\usepackage{caption}
\usepackage{soul}
\usepackage{fontawesome}
\usepackage{tabularray}
\usepackage[capitalize]{cleveref}
\crefname{section}{Sec.}{Secs.}
\Crefname{section}{Section}{Sections}
\Crefname{table}{Table}{Tables}
\crefname{table}{Tab.}{Tabs.}
\usepackage{academicons}
\usepackage{orcidlink}
\definecolor{aogreen}{rgb}{0.0, 0.5, 0.0}
\definecolor{applegreen}{rgb}{0.55, 0.71, 0.0}

\avssfinalcopy 

\def\avssPaperID{44} 

\ifavssfinal\fi
\begin{document}

\title{PGDS: Pose-Guidance Deep Supervision for Mitigating Clothes-Changing in Person Re-Identification}

\author{
Quoc-Huy Trinh$^{1}$, Nhat-Tan Bui$^{3}$, Dinh-Hieu Hoang$^{1, 2}$, Phuoc-Thao Vo Thi$^{1}$ \\ Hai-Dang Nguyen$^{1, 2}$, Debesh Jha$^{4}$, Ulas Bagci$^{4}$, Ngan Le$^{3}$, Minh-Triet Tran$^{1, 2}$ 
\vspace{1.7mm}\\
$^1$University of Science, Vietnam National University, Ho Chi Minh City, Vietnam  \\
$^2$John von Neumann Institute, Vietnam National University, Ho Chi Minh City, Vietnam \\
$^3$AICV Lab, University of Arkansas, Fayetteville, Arkansas, USA \\
$^4$Northwestern University, Chicago, Illinois, USA
}

\maketitle

\begin{abstract}

Person Re-Identification (Re-ID) task seeks to enhance the tracking of multiple individuals by surveillance cameras. It supports multimodal tasks, including text-based person retrieval and human matching. One of the most significant challenges faced in Re-ID is clothes-changing, where the same person may appear in different outfits. While previous methods have made notable progress in maintaining clothing data consistency and handling clothing change data, they still rely excessively on clothing information, which can limit performance due to the dynamic nature of human appearances. To mitigate this challenge, we propose the Pose-Guidance Deep Supervision (PGDS), an effective framework for learning pose guidance within the Re-ID task. It consists of three modules: a human encoder, a pose encoder, and a Pose-to-Human Projection module (PHP). Our framework guides the human encoder, i.e., the main re-identification model, with pose information from the pose encoder through multiple layers via the knowledge transfer mechanism from the PHP module, helping the human encoder learn body parts information without increasing computation resources in the inference stage. Through extensive experiments, our method surpasses the performance of current state-of-the-art methods, demonstrating its robustness and effectiveness for real-world applications. Our code is available at \url{https://github.com/huyquoctrinh/PGDS}.

\end{abstract}


\section{Introduction}
\label{sec:intro}
\noindent Person Re-Identification (Re-ID) involves tracking and distinguishing individuals across various cameras and viewpoints. This technique finds application in surveillance camera systems, enhancing security measures by enabling the search, tracking, and identifying unidentified individuals within specific areas or towns, thereby enhancing security measures. 



Despite notable advancements, several challenges persist in the Re-ID task, e.g. occlusion, weather conditions, and other external factors. In such problems, clothing-change presents itself as one of those paramount challenges, first highlighted by Xue et al. \cite{ccan} and Wu et al. \cite{wu2017robust}.
This challenge becomes particularly crucial in real-world scenarios where individuals always change their clothes. 

\begin{figure}[!t]
    \centering
    \includegraphics[width=0.4\textwidth]{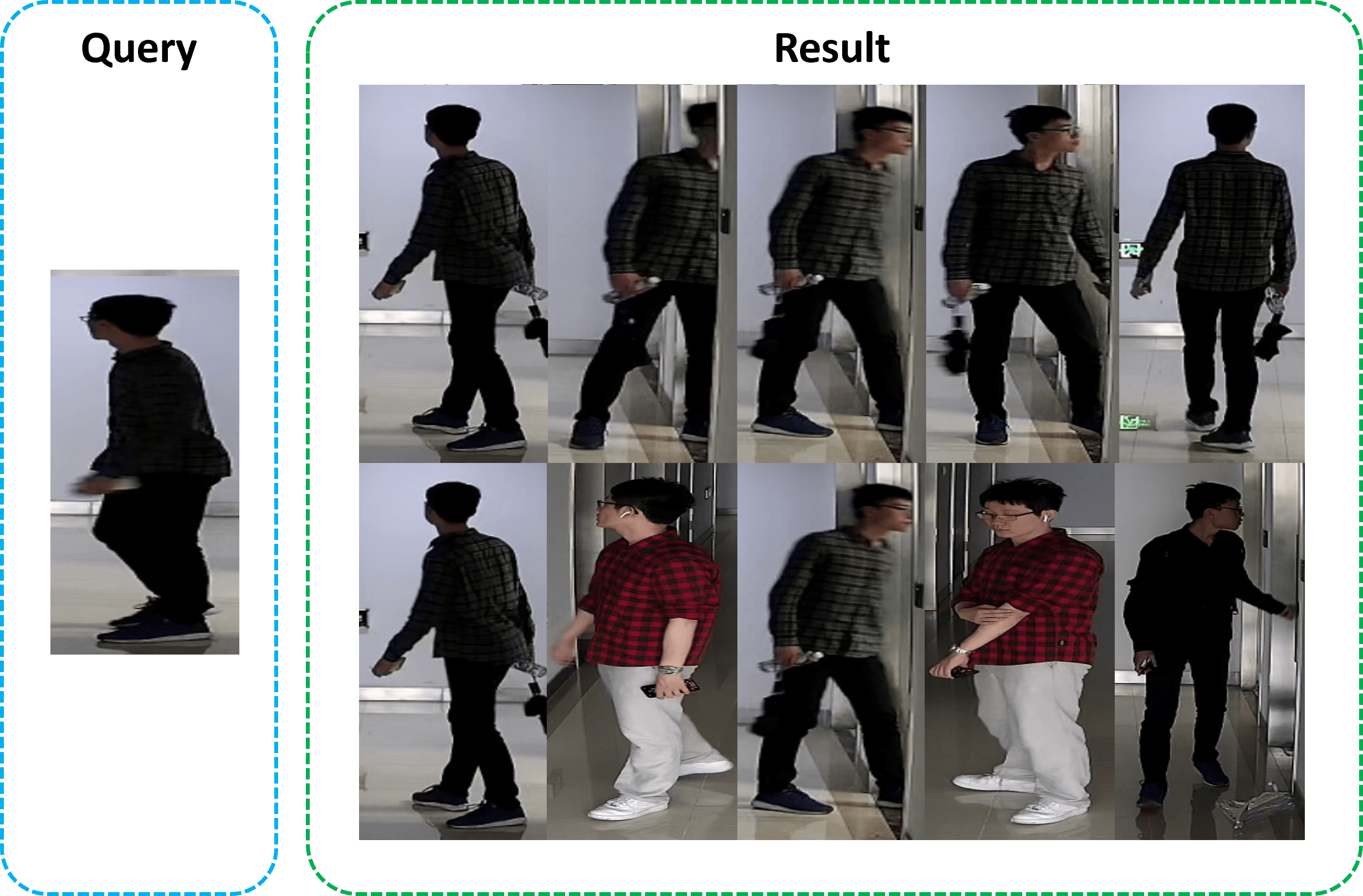}
    \caption{An example query retrieved by our framework under the clothes-changing scenario.}
    \label{fig:example}
\end{figure}


In recent years, researchers in the Re-ID domain have increasingly focused on mitigating the clothing-change problem. Various methods, such as VC-Clothes \cite{vcclothes}, LTCC \cite{qian2020long}, FSAM \cite{hong2021fine}, CAL \cite{gu2022clothes}, and GI-ReID \cite{jin2022cloth}, have been introduced to address this issue. The primary difficulty in
tackling this challenge arises from deep learning models relying heavily on clothing appearance to make predictions rather
than considering biometric traits (e.g., face, hair, pose, gait).
A growing body of literature, such as PGFL-KD \cite{pgflkd}, Pose-transfer \cite{Posedtri}, Pose-guided re-ID \cite{Pose-guidedre-ID}, and PSEAMA \cite{pseama} have demonstrated the effectiveness of pose information in tackling the clothes-chaning problem. However, these works often require complex modules to integrate body parts information into the Re-ID model, leading to an increase in computational cost, training time, and inference time.



As a result, we introduce \textbf{P}ose-\textbf{G}uidance \textbf{D}eep \textbf{S}upervision \textbf{(PGDS)}, a simple and innovative framework for the clothes-changing Re-ID task. Figure \ref{fig:example} indicates an example query result of our framework under the clothing-change scenario. In essence, our method builds upon three main modules: human encoder, pose encoder, and \textbf{P}ose-to-\textbf{H}uman \textbf{P}rojection module (PHP).
The human encoder, i.e., the main Re-ID model, efficiently adapts SOLIDER \cite{chen2023beyond}, a self-supervised learning framework for human-centric tasks, to extract the general human representations from the input images. At the same time, the pose encoder is taken from the pose estimation model OpenPose \cite{openpose} to generate meaningful global features for biometric identification, which includes the pose, body, foot, hand, and facial. Our PHP, which includes several projectors, operates between two encoders to transfer the pose information extracted from the pose encoder to the human encoder across various scales. In practice, we fine-tune the pre-trained human encoder and train our projectors while freezing the pose encoder since we only need the general human pose information from the pose encoder. By incorporating pose information into the multiple layers, the Re-ID model can effectively concentrate on the body parts information, which contain unique information crucial for individual identification. Furthermore, integrating the frozen pre-trained pose estimation model also avoids increasing the computational cost in the inference stage, thus benefiting real-world applications.
In summary, this paper’s main contributions are as follows: 
\begin{itemize}
    \item We explore the potential of transferring the human pose structure knowledge from the frozen pose estimation model into the main Re-ID network to overcome the clothes-changing problem. This simple strategy can be the baseline approach to alleviate the clothes-changing effect on the Re-ID task. 
    \item We introduce a Re-ID framework, termed PGDS, incorporating a frozen pre-trained pose estimation model to guide the human representation model across different scales.
    \item We conduct comprehensive ablation studies on multiple datasets to assess the effectiveness of our proposed PGDS framework.
\end{itemize}

The content of this paper is organized as follows. In Section \ref{sec:RelatedWork}, we briefly review existing methods for Clothes-changing Re-Identification problem and similar works that explore the utilization of pose information for Person Re-Identification. Then, we introduce our proposed method in Section \ref{sec:ProposedMethod}. Section \ref{sec:Experiment} presents the experiment setup, performance comparison, and ablation study. Finally, we conclude our work and suggest problems for future work in Section \ref{sec:Conclusion}.


\vspace{-2mm}
\section{Related Work}
\label{sec:RelatedWork}

\noindent\textbf{Clothes-changing Re-Identification:} Clothes-changing poses one of the most significant challenges in the Person Re-Identification task. The introduction of the LTCC dataset \cite{qian2020long} marked a significant development in addressing the clothes-changing problem. Subsequent research has yielded various methods to tackle this challenge. FSAM \cite{hong2021fine} aims to obtain coarse ID masks with structure-related details, incorporating ID-relevant information for discriminative structural feature extraction. CAL \cite{gu2022clothes} proposes a novel multi-positive-class classification loss to formulate multi-class adversarial learning. GI-ReID \cite{jin2022cloth} integrates clothes classification and a casual inference model to mitigate bias in clothing information. AIM \cite{yang2023good} defines a causality-based auto-intervention model to mitigate clothing bias for robust Clothes-changing Person Re-identification. FIR$e^{2}$ \cite{wang2023exploring} introduces fine-grained feature mining and a fine-grained attribute recomposition module to enhance the learning of robust features. Although previous works achieved impressive results on the clothes-changing Re-ID, it is worth noting that these methods lack focusing on the crucial parts of the body, which is the salient information to distinguish a specific person. In contrast, our work focus on prioritizing body parts information by efficiently leverage the pose information from frozen pre-trained pose estimation model to guide the main Re-ID model.


\noindent
\textbf{Pose Guide for Clothes-changing Re-Identification:} Pose is crucial information utilized to guide Re-Identification models. Several methods have been proposed based on this approach. Among the most notable ones are ABDNet \cite{abd}, PGR \cite{pgr}, Gated Fusion \cite{gatedfusion}, and Pitr \cite{pitr}. These methods follow the concept of integrating modules to fuse features extracted from pose heatmaps to the Re-ID models. More recently, PGFL-KD \cite{pgflkd}, PFD \cite{tao}, Pose-guided re-ID \cite{Pose-guidedre-ID}, and PSEAMA \cite{pseama} have leveraged pose guidance through part matching via global features. These methods demonstrate the effectiveness of pose information in guiding Re-ID models. However, the primary issue noted in these approaches is the substantial computational load imposed by the fusion modules, which are used to incorporate pose information into the main model.
Differently from these previous works, PGDS integrates the pre-trained pose estimation model during the training phase, which remains frozen, to guide the main Re-ID model with pose information. This approach prevents an increase in computational expenses during the inference stage.

\section{Proposed PGDS}
\label{sec:ProposedMethod}
\begin{figure*}[!t]
    \centering
    \includegraphics[width=0.8\textwidth]{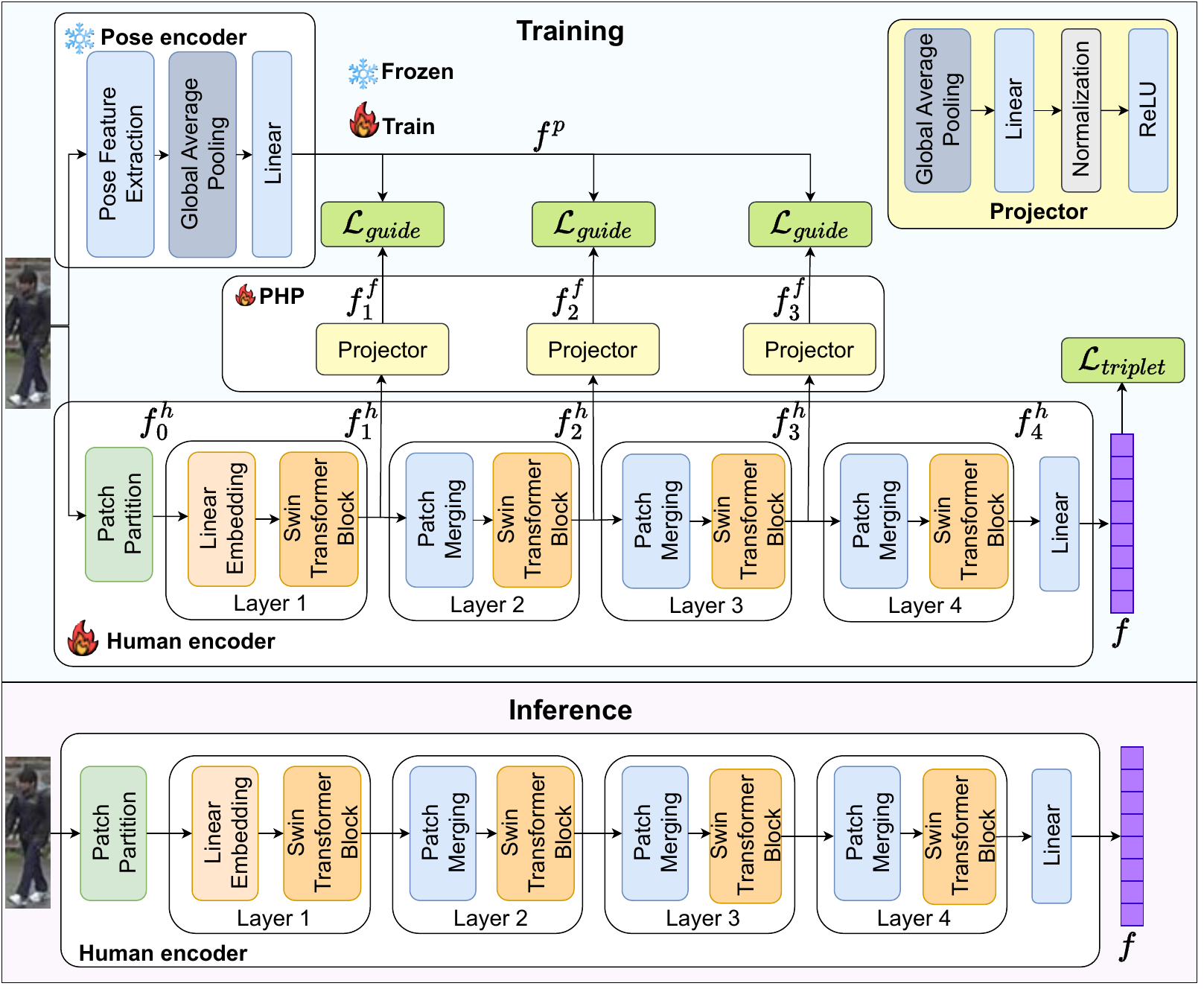}
    \caption{Overall framework of proposed PGDS including three modules: a human encoder, a pose encoder, and a pose-to-human projection module (PHP). The pose encoder module utilizes a frozen pre-trained model while we fine-tune a pre-trained human-centric model for the human encoder module. Our PHP transfers pose knowledge from the pose encoder module to the human encoder module through multiple projectors and guide loss $\mathcal{L}_{guide}$. $H, W,$ and $C$ denote the height, width, and channel, respectively. $\mathcal{L}_{triplet}$ is the triplet loss~\cite{hoffer2015deep} which acquires person-centric representations.}
    \label{fig:vachi}

\end{figure*}

\noindent In this section, we introduce our PGDS framework, which consists of three primary modules: the human encoder, the pose encoder, and the pose-to-human projection module (PHP), as illustrated in Figure~\ref{fig:vachi}.

The human encoder takes in the image, learns each person's feature representations and unique human appearance information, and then returns the image embedding. Meanwhile, the pose encoder (i.e., from a frozen pre-trained pose estimation model) is used to extract the human pose information from the input image and packs this information into a pose heatmap. In the training phase, the PHP encourages the early layers of the human encoder to attain the human representation with additional pose information through a knowledge distillation mechanism. When inferring, only feature representations from the human encoder are used to accomplish the re-identification task. In other words, we leverage a frozen encoder pre-trained for pose estimation as a \textbf{teacher} to compel the \textbf{student} human encoder to retain pose information into the image embedding. We detail each component in the following subsections to provide more in-depth information.

\subsection{Pose Encoder}



\noindent As mentioned, the proposed PGS aims to transfer the gait, pose, and posture information, which are unique characteristics of each person, to consolidate the model's capacity to identify the same person wearing different clothes. Thus, we propose a simple strategy that exploits a frozen pre-trained pose estimator to enhance the Re-ID model's ability to concentrate on body part information instead of cloth appearance.

Our pose encoder is derived from the OpenPose \cite{openpose}, a popular 2D pose estimation model. We hypothesize that the feature representation from OpenPose sufficiently and efficiently covers the pose-related information for our framework since we only want to get the general posture information. As a consequence, we keep the pose encoder unchanged so that we can minimize the number of parameters during training.

The pose encoder takes the input human image to generate the confidence map (i.e. pose feature), which highlights parts of the human, including ``human body, hand, facial, and foot keypoints'' \cite{openpose}. The confidence map is then passed through the global average pooling and fully connected layer to produce the meaningful feature maps $f^{p}\in\mathbb{R}^{1 \times 768}$ that are associated with the overall posture information.

\subsection{Human Encoder}
\noindent Our human encoder is derived from SOLIDER \cite{chen2023beyond}, a human-centric self-supervised learning framework. It is important to note that the SOLIDER framework is based on the teacher-student knowledge distillation approach. Thus, it can adapt to a wide range of downstream tasks with different requisitions of semantic information and appearance information. In addition to the teacher and student networks, SOLIDER has a novel semantic controller that can modify the ratio of semantic over appearance information encoded in the feature. 

Following the original paper \cite{chen2023beyond}, we use SOLIDER's student network as our human encoder, accompanied by the frozen semantic controller with the ratio $\lambda_{human}=0.2$.
More precisely, the backbone of the human encoder is the Swin Transformer \cite{liu2021swin}, which consists of four main layers. First, the human encoder will encode the image $X\in\mathbb{R}^{W \times H \times 3}$ into five scale features $f_{i}^{h} \in \mathbb{R}^{\frac{W}{2^{i+1}} \times \frac{H}{2^{i+1}} \times C_{i}}$ where $C_{i} \in \{48, 96, 192, 384, 768\}$, and $i \in \{0,1,2,3,4\}$. The final scale feature $f^h_4$ undergoes the linear layer to generate the final human embedding, denoted as $f\in\mathbb{R}^{1 \times 768}$. Each human embedding will be utilized to retrieve each individual. We train this human encoder in the contrastive learning manner with the triplet loss so that the model can discriminate between different people. This image embedding will also be aligned with the pose feature $f^p$ by the guide loss function \cite{hoffer2015deep}, which we will detail in the following subsection \ref{subsection-obj-func}. 

\subsection{Pose-to-Human Projection Module (PHP)} 
\noindent The primary objective of PHP is to effectively and robustly transfer posture knowledge from the pose encoder to the human encoder. To enhance the consistency of the human encoder throughout all its stages, PHP incorporates three projectors corresponding to the three intermediate layers of the human encoder. At each stage, denoted as $i$, the projector begins with the input feature from the human encoder, referred to as $f^h_i$. It then extracts implicit human representations, denoted as $f^f_i$, through a series of operations, including global average pooling, linear projection, batch normalization, and applying the non-linear ReLU activation function. 



It is important to note that the human encoder outputs feature vectors that are more informative and semantically richer compared to the pose encoder, which only contains human pose information. From this understanding, it becomes apparent that directly guiding the features from the human encoder with pose features might result in the loss of other crucial information embedded in the human features. Therefore, we design an indirect transfer knowledge mechanism called a projector. This mechanism enables the human encoder to learn pose information without compromising other essential information. Intuitively, the projector acts as a non-linear, lightweight extractor. It unveils human-specific information within the higher-dimensional image representation of the human encoder and maps it to the lower-dimensional pose heatmap space of the pose encoder. Consequently, the resulting features at each stage, denoted as $f^f_i$, are incorporated into the guide loss $\mathcal{L}_{guide}$ to align them with the pose feature $f^p$ obtained from the pose encoder.

\subsection{Objective Functions}
\label{subsection-obj-func}
\noindent Our framework is trained end-to-end with a combined loss, defined as in Equation \ref{equa:combined}.
\begin{equation}
    \centering
    \mathcal{L}_{combined}= \mathcal{L}_{triplet} + \lambda \mathcal{L}_{guide}
    \label{equa:combined}
\end{equation}
where $\mathcal{L}_{triplet}$ is the triplet loss \cite{hoffer2015deep} to learn person-specific representations, and $\mathcal{L}_{guide}$ is the guide loss to transfer the pose knowledge from pose encoder to human encoder. The hyperparameter $\lambda$, which we set to 0.8, controls the contribution and effect of the guide loss on the whole system. See section \ref{lambda} for the ablation study on the $\lambda$ values.

\noindent\textbf{Triplet Loss:} to learn a discriminative embedding in the feature space (i.e., encouraging representations of the same person to be similar to one another), we adapt the popular deep metric learning triplet loss \cite{hoffer2015deep}, which is defined as follows:
\begin{equation}
\label{equa:triplet}
    \mathcal{L}_{triplet}= \sum^{N}_{i=1}[||f_{i}^{an} -
    f_{i}^{po}||^{2}_{2} -
    ||f_{i}^{an} - f^{ne}_{i}||^{2}_{2} + \alpha]
\end{equation}
where $N$ corresponds to the number of data samples in a batch, $i$ denotes the index of the sample, $f$ stands for the human embedding from the human encoder, and ${an, po, ne}$ symbols individually refer to the anchor, positive, and negative instances. We employ batch hard-triplet mining \cite{hardmining} for the triplet selection. The parameter $\alpha$ signifies the margin, dictating the separation between embeddings. We set $\alpha$ equal to 0.2 in our work. 

\noindent\textbf{Guide Loss:} 
To encourage the model to learn semantically consistent and clustered feature representations, we utilize KL-divergence loss and the cross-entropy defined on the class-probability vectors $p^p$ and $p^f$ of pose embedding $f^{p}$ and human representation $f^{f}$:
\begin{equation}
    \mathcal{L}_{KL}(p^p, p^f) = y\mathcal{L}_{sim}(p^p, p^f) + (1-y)\mathcal{L}_{dis}(p^p, p^f)
\end{equation}
If $p^p$ and $p^f$ share the same human ID, $y = 1$ and 
\begin{equation}
\small
\mathcal{L}_{KL}(p^p, p^f) = \mathcal{L}_{sim}(p^p, p^f)  = KL(p^p|p^f) + KL(p^f|p^p)
\end{equation}
Otherwise, $y = 0$ and 
\begin{equation}
\small
\begin{split}
\mathcal{L}_{KL}(p^p, p^f) = \mathcal{L}_{dis}(p^p, p^f)  = max\left(0,\left(m-KL(p^p|p^f)\right)\right)\\ + max\left(0,\left(m-KL(p^f|p^p)\right)\right)
\end{split}
\end{equation}
where $m$ is user-specified margin and $m=2$. The KL-divergence is given by $KL(p|q) = \sum_{k=1}^K(p^k log(\frac{p^k}{q^k}))$

Corresponding to three layers in PHP, the guide loss at each layer $i^{th}$ is computed as $\mathcal{L}_{KL}(p^p, p^f_i)$, where $p^p$ and $p^f_i$ are class-probability vectors of pose embedding $f^{p}$ and human representation $f^{f}_i$.

During the training phase, at each stage, the primary objective of this loss function is to minimize the disparity between the distributions originating from pose embeddings $f^p$ and those obtained from human embeddings $f^f_i$. This process helps determine the optimal weights for refining human appearance representations by incorporating pose knowledge. Consequently, the model becomes proficient at capturing appearance and pose-related information. Since we aim to introduce a simple framework for transferring biometric knowledge (i.e., pose) to the Re-ID task, we have not conducted an exhaustive search for the best loss functions. This allows other researchers to explore the potential of alternative loss functions for this task.

\section{Experiments}
\label{sec:Experiment}
\subsection{Experimental Setup}

\noindent\textbf{Datasets:} 
To assess the effectiveness of the proposed PGDS, we conduct evaluations on five datasets that previous methods have widely used. These datasets cover a wide range of scenarios and closely resemble real-world conditions. \underline{Market-1501} \cite{market1501} contains 1,501 identities observed from 6 camera viewpoints, including 12,936 training images of 751 identities, 19,732 gallery images, and 2,228 queries.
\underline{Duke-MTMC} \cite{dukemtmc} comprises 34,183 images of 1,404 identities from eight cameras. It contains 16,522 training images, 17,661 gallery images, and 2,228 queries.
\underline{CuHK03} \cite{cuhk032} contains 14,097 pictures of 1,467 identities. Additionally, the dataset offers 20 sets for training and testing, where 100 identities are reserved for testing in each set while the remaining identities are used for training. 
\underline{LTCC} \cite{qian2020long} contains 17,119 person images of 152 identities. The training set in this data includes 9,576 images with 77 identities (46 clothes-changing IDs and 31 clothes-consistent IDs). The testing set includes 493 query images and 7,050 test images with 75 identities (45 clothes-changing IDs + 30 clothes-consistent IDs). This dataset includes two types of testing: \textit{Standard} (test cases including clothes-changing and clothes-consistent), and \textit{CC} (test cases include only clothes-changing IDs).
\underline{VC-Clothes} \cite{vcclothes} comprises synthesized human images. It encompasses 19,060 images of 512 unique identities in four scenes. In the training phase, 9449 images of 256 identities are employed, leaving the remaining 9611 images of 256 identities for the testing phase.

\noindent\textbf{Implementation Detail: }
During the training and testing phase, all images are resized to dimensions of $384 \times 128$. Our training and testing for the clothes-consistent follow the setting of Bag of Trick \cite{luo2019bag}, while in the clothes-changing testing, we follow the setup of LTCC \cite{qian2020long}. The Pytorch framework is used, and all experiments are done on the Tesla A100 40GB GPU. The batch size for the training and testing is 64; we also use the learning rate scheduler to support our training experiments with a base learning rate of 8e-4, and the temperature value equals 2. We choose the AdamW\cite{loshchilov2017decoupled} as our optimizer, and the best weight has been gotten after 250 epochs. For data augmentation, we employ random horizontal flip and random erasing.

\noindent\textbf{Baselines: }
Based on two kinds of Re-ID problems, we select different baselines in two scenarios as follows.
For \underline{clothes-changing} setting, we evaluate our method on LTCC-Standard, LTCC-CC and VC-Clothes datasets. We compare our method with other state-of-the-art methods (i.e., LTCC \cite{qian2020long}, FSAM \cite{hong2021fine}, CAL \cite{gu2022clothes}, GI-ReID \cite{jin2022cloth}, AIM \cite{yang2023good}, and FIRe$^{2}$ \cite{wang2023exploring}). 
For \underline{clothes-consistent} setting, we evaluate our method on Market-1501, Duke-MTMC and CuHK03 datasets. We compare our method with several state-of-the-art methods on the Re-ID task, including Bag of Trick \cite{luo2019bag}, ABDNet \cite{abd},  Auto-ReID+ \cite{autoreid+}, TransReID \cite{he2021transreid}, PGFL-KD \cite{pgflkd}, AML \cite{aml}, PFD \cite{tao}, and SOLIDER \cite{chen2023beyond}.
To further assess the robustness of our model, we conduct the \underline{cross-domain} setting in two scenarios: the Duke to Market ($D \longrightarrow M$) and the Market to Duke ($M \longrightarrow D$). We compare our method against several existing methods, e.g., SSL \cite{luo2021self}, ATNet \cite{atnet}, UDAP \cite{udap}, and SOLIDER \cite{chen2023beyond}.

\noindent\textbf{Evaluation Metrics: }
For a fair comparison, we employ two metrics, i.e., mean Average Precision (mAP), and Rank-1 accuracy (R1). mAP is calculated via the area under the precision-recall curve while R1 is defined by how many samples are correct in the top-1 prediction.

\subsection{Performance Comparisons}
\label{sec:Result}
\noindent The results of the clothes-changing setting, the clothes-consistent setting, and the cross-domain setting are shown in Table \ref{table:ltcc}, Table \ref{table:same}, and the Table \ref{table:crossdomain}, respectively.
\begin{table}[!h]
\centering
\caption{Quantitative results on the clothes-changing datasets. The highest scores are shown in \textbf{bold}.}
\resizebox{\columnwidth}{!}{%
\begin{tabular}{l|cc|cc|cc}
\toprule
\multirow{2}{*}{\textbf{Methods}}  &  \multicolumn{2}{c|}{\textbf{LTCC-Standard}} & \multicolumn{2}{c|}{\textbf{LTCC-CC}} & \multicolumn{2}{c}{\textbf{VC-Clothes}} \\ \cline{2-7}
 & \textbf{mAP}$\uparrow$   & \textbf{R1}$\uparrow$    & \textbf{mAP}$\uparrow$   & \textbf{R1}$\uparrow$ & \textbf{mAP}$\uparrow$   & \textbf{R1}$\uparrow$  \\ \hline
LTCC \cite{qian2020long} &  34.3 & 71.4   & 11.7  & 25.2 & $--$ & $--$\\
FSAM \cite{hong2021fine} & 35.4 & 73.2 & 16.2 & 38.5 & 78.9 & 78.6 \\
CAL \cite{gu2022clothes} & 40.8 & 74.2 & 18.0 & 40.1 & $--$ & $--$  \\
GI-ReID \cite{jin2022cloth}  & 29.4 & 63.2 & 10.4 & 23.7 & 59.0 & 63.7 \\
AIM \cite{yang2023good}  & 41.1 & {76.3} & {19.1} & 40.6 & 74.1 & 73.7\\
FIRe$^{2}$ \cite{wang2023exploring} & 39.9 & 75.9 & 19.1 & 44.6 & $--$ & $--$ \\
\hline
\textbf{PGDS (Ours)} & \textbf{43.0} & \textbf{77.5} & \textbf{26.7} & \textbf{49.1} &  \textbf{84.6} & \textbf{92.5}\\
\bottomrule
\end{tabular}}
\label{table:ltcc}
\end{table} 

\noindent\textbf{Clothes-changing:} From Table \ref{table:ltcc}, it is evident that on the LTCC-Standard dataset, our model showcases an improvement of over $+1.9\%$ in mAP and $+1.2\%$ in the R1 metric compared to the second-best method, AIM \cite{yang2023good}. In the sole clothes-changing testing scenario (LTCC-CC dataset), our method outperforms the second-best (FIRe$^2$) with a $+7.6\%$ improvement in mAP and a substantial margin of $+4.5\%$ in the R1 metric. As for the VC-Clothes dataset, our method achieves a notable improvement of $+5.7\%$ in mAP and $+13.9\%$ in the R1 metric compared to the second-best method, FSAM. These results underscore the efficacy of our method in addressing the clothes-changing problem by prioritizing other body parts over clothing information.

\begin{table}[!h]
\centering
\caption{Quantitative results on the clothes-consistent datasets. The highest scores are shown in \textbf{bold}.}

\resizebox{\columnwidth}{!}{%
\begin{tabular}{l|cc|cc|cc}
\toprule
\multirow{2}{*}{\textbf{Methods}} &  \multicolumn{2}{c|}{\textbf{Market-1501}} & \multicolumn{2}{c|}{\textbf{Duke-MTMC}} & \multicolumn{2}{c}{\textbf{CuHK03}} \\\cline{2-7}
 &  \textbf{mAP}$\uparrow$  & \textbf{R1}$\uparrow$    & \textbf{mAP}$\uparrow$   & \textbf{R1}$\uparrow$  & \textbf{mAP}$\uparrow$   & \textbf{R1}$\uparrow$ \\ \hline
Bag of Trick \cite{luo2019bag} & 94.2 & 95.4   & 89.1  & 90.3 & 56.6  & 58.8\\
ABDNet \cite{abd}  &    88.3  &  95.6   & 78.6  & 89.0 & $--$  & $--$\\
Auto-ReID+ \cite{autoreid+} &  88.2 & 95.8 & 80.1 & 90.1 & 74.2 & 78.1\\
PGFL-KD \cite{pgflkd} &  87.2 & 95.3 & 79.5 & 89.6 & $--$ & $--$ \\
AML \cite{aml} &  89.5 & 95.7 & 81.7 & 91.1 & 82.3 & 85.6 \\
PFD \cite{tao} &   89.7 & 95.5 & 82.2 & 90.6 & $--$ & $--$ \\
SOLIDER \cite{chen2023beyond}  & 95.3 & 96.6 & 86.1 & 89.4 & 71.6 & 67.4\\
\hline
\textbf{PGDS (Our)} & \textbf{95.4} & \textbf{96.9} & \textbf{91.4} & \textbf{92.6} & \textbf{89.7} & \textbf{87.9}\\

\bottomrule
\end{tabular}}
\label{table:same}
\end{table}

\noindent\textbf{Clothes-consistent:} 
From Table \ref{table:same}, it is evident that our method leads to an improvement of $+0.1\%$ in the mAP metric and $+0.3\%$ in the R1 metric on the Market-1501 dataset when compared to the second-best method, SOLIDER \cite{chen2023beyond}. Regarding the Duke-MTMC dataset, our method achieves a better mAP metric by $+2.3\%$ and an improvement of $+2.3\%$ in the R1 metric compared to the second-best method, Bag of Tricks \cite{luo2019bag}. In the CuHK03 dataset, our method surpasses the second-best method, AML \cite{aml}, by a margin of $+7.4\%$ in the mAP metric and $+2.3\%$ in the R1 metric.  These results demonstrate that our method remains competitive with state-of-the-art approaches across all three datasets, emphasizing the efficacy of leveraging pose information for distinguishing individuals. Such outcomes are significant as they highlight the potential robustness of our approach for future studies.



\noindent\textbf{Cross-domain testing:}  From Table \ref{table:crossdomain}, in the $D \longrightarrow M$ setting, our method surpasses the second best method by margins of $+9.9\%$ and $+1.2\%$ in terms of mAP and R1, respectively. While in the $M \longrightarrow D$ scenario, we exceed the performance by $+3.2\%$ and $+3.7\%$ in mAP and R1 compared to the second best method. These results highlight the robustness of our framework in the Re-ID task.


\begin{table}[!h]
\centering
\caption{Results of the cross-domain testing. D, M represent the Duke-MTMC and Market-1501 datasets. The highest scores are shown in \textbf{bold}.}
\resizebox{.8\columnwidth}{!}{%
\begin{tabular}{l|cc|cc}
\toprule
\multirow{2}{*}{\textbf{Methods}}  &  \multicolumn{2}{c|}{\textbf{D $\longrightarrow$ M}} & \multicolumn{2}{c}{\textbf{M $\longrightarrow$ D}} \\ \cline{2-5}
 & \textbf{mAP}$\uparrow$   & \textbf{R1}$\uparrow$    & \textbf{mAP}$\uparrow$   & \textbf{R1}$\uparrow$  \\ \hline
SSL \cite{luo2021self}  & 	 	37.8 & 	71.7  & 	28.6 & 	52.5 \\
ATNet \cite{atnet} & 	 25.6 & 55.7  & 	24.9 & 	45.1 \\
UDAP \cite{udap} & 	 	{53.7} & 	75.8 &	49.0 & 	68.4 \\
SOLIDER \cite{chen2023beyond} & 	 	50.3 & 	59.9 &  50.3 & 	59.9 \\
\hline
\textbf{PGDS (Our)} & \textbf{63.6} & \textbf{77.0} & \textbf{53.5} & \textbf{72.1}\\
\bottomrule
\end{tabular}}
\label{table:crossdomain}
\end{table}

\subsection{Ablation Study}
\label{subsec-ablation}
\noindent To evaluate the effectiveness of the PGDS framework, we conducted four ablation studies. The first experiment assesses the effectiveness of pose guidance by the PHP module across multiple scales in the clothes-consistent problem, compared to the baseline human encoder (i.e., SOLIDER \cite{chen2023beyond}). The second experiment verifies the PHP module's effectiveness across multiple scales in the clothes-changing problem.
The third experiment explores the impact of the parameter $\lambda$, which controls the contribution of the guide loss in the total loss, on knowledge transfer. 

\noindent\textbf{Effect of PHP across multiple scales in clothes-consistent problem:}  Table~\ref{table:ablation-consistent} shows that with our proposed PHP module applying to numerous stages, the results are improved over the baseline, even with a relatively small number of added trainable parameters. Since our framework only uses the human encoder in the inference stage, thus the FLOPs value remains the same compared with the baseline.

\begin{table}[H]
\centering
\caption{Impact of PHP in clothes-consistent problem compared with baseline in Duke-MTMC dataset. Our method, represented by three distinct versions, i.e., PGDS-1, PGDS-2, and PGDS-3, incorporates one to three PHP projectors for transferring posture knowledge.}
\resizebox{\columnwidth}{!}{%
\begin{tabular}{l|cc|cc}
\toprule
\multirow{2}{*}{\textbf{Methods}} & \multicolumn{2}{c|}{\textbf{Computational Cost}} & \multicolumn{2}{c}{\textbf{Metrics}}\\ \cline{2-5}
 &  \textbf{Params(M)}$\downarrow$ & \textbf{FLOPs(G)}$\downarrow$ & \textbf{mAP}$\uparrow$   & \textbf{R1}$\uparrow$  \\ \hline
SOLIDER \cite{chen2023beyond} & 27.51 & 5.54 & 86.1 & 89.4\\
PGDS-1 & 28.13(\textcolor{applegreen}\faCaretUp 0.62) & 5.54 & 88.6(\textcolor{aogreen}\faCaretUp 2.5) & 90.9(\textcolor{aogreen}\faCaretUp 1.5) \\ 
PGDS-2 & 28.28(\textcolor{applegreen}\faCaretUp 0.77) & 5.54 & 89.6(\textcolor{aogreen}\faCaretUp 3.5) & 91.4(\textcolor{aogreen}\faCaretUp 2.0) \\
PGDS-3  & 28.57(\textcolor{applegreen}\faCaretUp 1.06)  & 5.54 & 91.4(\textcolor{aogreen}\faCaretUp 5.3) & 92.6(\textcolor{aogreen}\faCaretUp 3.2)\\

\bottomrule
\end{tabular}}
\label{table:ablation-consistent}
\end{table}
\noindent\textbf{Effect of PHP across multiple scales in clothes-changing problem:} The results are summarized in Table~\ref{table:ablation-changing} indicate that when integrating the projector of PHP into multiple stages, the Re-ID framework can make more robust prediction since it prioritizes the pose information more than the clothes.
\begin{table}[H]
\centering
\caption{Impact of PHP in clothes-changing problem.}
\resizebox{\columnwidth}{!}{%
\begin{tabular}{l|cc|cc}
\toprule
\multirow{2}{*}{\textbf{Settings} } &  \multicolumn{2}{c|}{\textbf{LTCC-CC}} & \multicolumn{2}{c}{\textbf{VC-Clothes}} \\
\cline{2-5}
 & \textbf{mAP}$\uparrow$   & \textbf{R1}$\uparrow$    & \textbf{mAP}$\uparrow$   & \textbf{R1}$\uparrow$  \\ \hline

PGDS-1 & 23.5 & 45.1 & 83.3 & 91.3\\ 

PGDS-2 & 24.4(\textcolor{aogreen}\faCaretUp 0.9)& 45.6(\textcolor{aogreen}\faCaretUp 0.5)& 83.4(\textcolor{aogreen}\faCaretUp 0.1) & 91.4(\textcolor{aogreen}\faCaretUp 0.1)\\

PGDS-3 & 26.7(\textcolor{aogreen}\faCaretUp 3.2)& 49.1(\textcolor{aogreen}\faCaretUp 4.0)& 84.6(\textcolor{aogreen}\faCaretUp 1.3) & 92.5(\textcolor{aogreen}\faCaretUp 1.2)\\

\bottomrule
\end{tabular}}
\label{table:ablation-changing}
\end{table}

\noindent\textbf{Impact of $\lambda$ for knowledge transfer:} 
\label{lambda} As illustrated in Table~\ref{table:ablationlambda}, the results indicate that the model's performance first improves and then declines when $\lambda$ equals 1.0. This phenomenon occurs because a low $\lambda$ limits the model's ability to adapt to pose knowledge, while setting $\lambda$ to 1.0 can make it challenging for the model to optimize both the triplet loss and guide loss simultaneously.
\begin{table}[H]
\centering
\caption{Ablation of the $\lambda$ parameters on LTCC-CC dataset.}
\resizebox{\columnwidth}{!}{%
\begin{tabular}{l|ccccc}
\toprule
\textbf{Accuracy} &  $\lambda$ = \textbf{0.2}  & $\lambda$ = \textbf{0.4} & $\lambda$ = \textbf{0.6} & $\lambda$ = \textbf{0.8} & $\lambda$ = \textbf{1.0} \\ \hline

\textbf{mAP}$\uparrow$ &25.2 & 25.6 & 25.9 & \textbf{26.7} & 26.2\\

\textbf{R1}$\uparrow$  & 45.8 & 48.6 & 48.9 & \textbf{49.1} & 45.8\\

\bottomrule
\end{tabular}}
\label{table:ablationlambda}
\end{table}

\subsection{Feature Map Visualization}

\begin{figure}[ht]
    \centering
    \includegraphics[width=\linewidth]{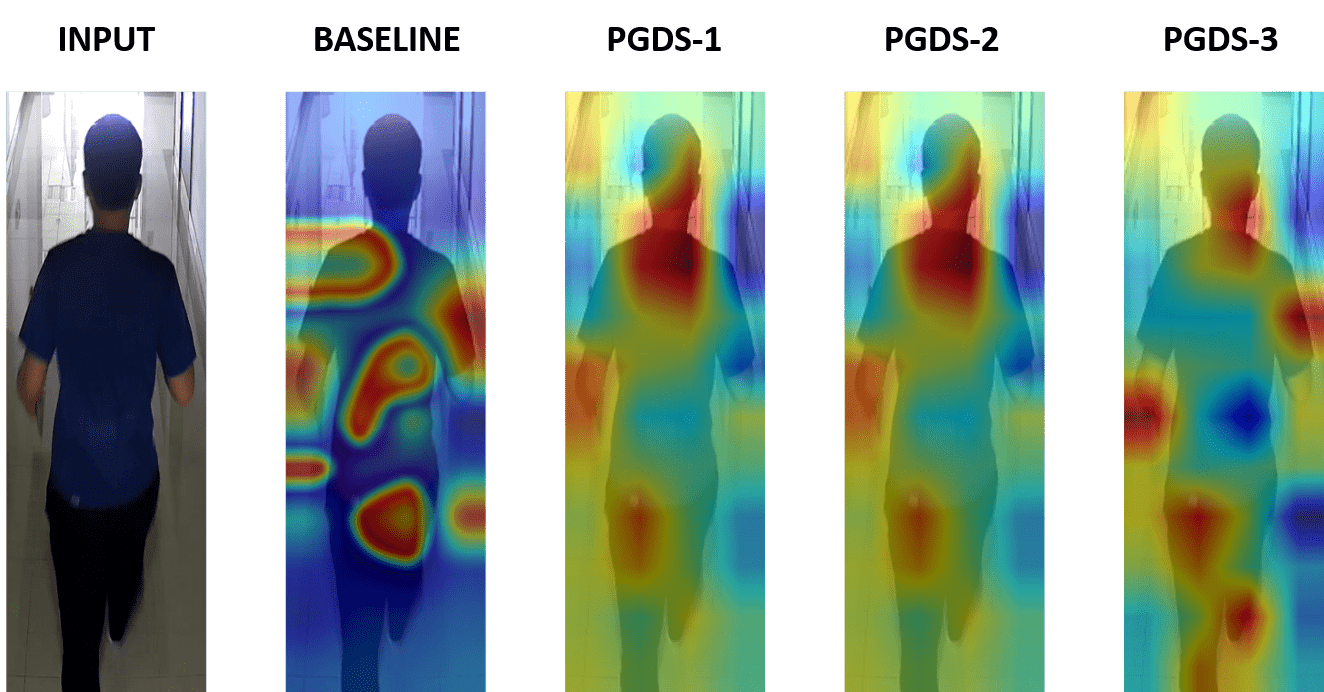}
    \caption{Heatmap visualization compared with baseline to understand the behavior of our framework. The baseline is the SOLIDER \cite{chen2023beyond}.}
    \label{fig:vis}
\end{figure}

\noindent To gain deeper insights into the behavior of our framework, we conducted heatmap visualization, as depicted in Figure~\ref{fig:vis}. This visualization readily highlights that the resulting heatmaps from the three versions (PGDS-1, PGDS-2, PGDS-3) primarily concentrate on the human's head and various body parts. In simpler terms, the synergy of the human encoder, pose encoder, and the PHP module equips the model with the ability to accurately prioritize the human body, reducing its dependence on clothing variations.

\section{Conclusion}
\label{sec:Conclusion}

\noindent In conclusion, this paper introduces the concept of Pose-Guidance Deep Supervision (PGDS), which transfers the pose knowledge into the main Re-ID model at multiple scales for robust Re-ID in scenarios involving both clothes-consistent and clothes-changing. Our study demonstrates the effectiveness of this approach in enabling the model to learn robust features, as evidenced by its competitive performance compared to current state-of-the-art methods. This simple approach is a promising candidate for real-world camera surveillance applications and provides a solid foundation for future studies to build more robust models. We encourage fellow researchers to explore various methods to improve this framework and address cloth-changing problems through the approach that integrates pose information.

\noindent
\textbf{Acknowledgement.}
Dinh-Hieu Hoang is funded by Vingroup Joint Stock Company and supported by the Domestic Master/ PhD Scholarship Programme of Vingroup Innovation Foundation (VINIF), Vingroup Big Data Institute (VINBIGDATA), code VINIF.2022.ThS.JVN.04. 
Debesh Jha and Ulas Bagci are supported by the NIH funding: R01-CA246704, R01-CA240639, U01-DK127384-02S1, and U01-CA268808.

{\footnotesize
\bibliographystyle{ieee}
\bibliography{egbib}
}

\end{document}